\documentclass{article} 
\usepackage[preprint]{colm2025_conference}

\usepackage{microtype}
\usepackage{hyperref}
\usepackage{url}
\usepackage{booktabs}

\usepackage{lineno}

\usepackage[T1]{fontenc}
\usepackage[utf8]{inputenc}
\usepackage{microtype}
\usepackage{inconsolata}
\usepackage{graphicx}
\usepackage{tikz}
\usepackage{subcaption}
\usepackage{pgfplots}
    \pgfplotsset{
        compat=1.3,
    }
\usepackage{booktabs}
\usepackage{amsmath}
\usepackage{verbatim}

\definecolor{darkblue}{rgb}{0, 0, 0.5}
\hypersetup{colorlinks=true, citecolor=darkblue, linkcolor=darkblue, urlcolor=darkblue}

\title{Factored Agents: Decoupling In-Context Learning and Memorization for Robust Tool Use}

\author{Nicholas Roth\thanks{Corresponding author: nicholasroth@google.com} \\
Google \\
\texttt{nicholasroth@google.com} \\
\And
Christopher Hidey \\
Google \\
\And
Lucas Spangher \\
Google, MIT \\
\AND
William F. Arnold \\
Google, KAIST \\
\And
Chang Ye \\
Google \\
\And
Nick Masiewicki \\
Google \\
\AND
Jinoo Baek \\
Google \\
\And
Peter Grabowski \\
Google, UC Berkeley \\
\And
Eugene Ie \\
Google
}

\begin{document}

\ifcolmsubmission
\linenumbers
\fi

\maketitle

\begin{abstract}
In this paper, we propose a novel factored agent architecture designed to overcome the limitations of traditional single-agent systems in agentic AI. Our approach decomposes the agent into two specialized components: (1) a large language model (LLM) that serves as a high level planner and in-context learner, which may use dynamically available information in user prompts, (2) a smaller language model which acts as a memorizer of tool format and output. This decoupling addresses prevalent issues in monolithic designs, including malformed, missing, and hallucinated API fields, as well as suboptimal planning in dynamic environments. Empirical evaluations demonstrate that our factored architecture significantly improves planning accuracy and error resilience, while elucidating the inherent trade-off between in-context learning and static memorization. These findings suggest that a factored approach is a promising pathway for developing more robust and adaptable agentic AI systems.
\end{abstract}

\section{Introduction}

Recent advances in Artificial Intelligence (AI) have brought forth the promise of ``Agentic AI'' systems endowed with a degree of autonomy and goal-directed behavior that could fundamentally transform computing and human–computer interactions. Agentic AI can be broadly defined as a class of models that can competently and reliably interact with external APIs, which they may use to interact with tools and systems. For our purposes, we note that they may synthesize both learned knowledge and contextual cues to achieve their objectives. Such systems hold the potential to not only perform complex computations but also to adapt dynamically to new tasks and environments \citep{acharya2025agentic}. 

 In its most common instantiation, agentic AI is generally proposed as a single-agent design. This baseline approach involves a monolithic Large Language Model (LLM) responsible for all aspects of planning, memory, and action execution. Such a system may be effective in a static environment when the format of the APIs that interact with external systems are included in its training or fine-tuning set. 
 
 However, in a large, dynamic, and real-world system, a single agent is ill-suited. Specifically, the single-agent framework is susceptible to several types of errors: those at the small scale of producing correct structured API output, and those at the larger scale of deciding what to do. In the small scale, we note that many of the following errors may come from \textbf{parsing errors}:
 
 \begin{enumerate}
     \item Malformed Fields: the LLM may misspell or misunderstand the input type required from it
     \item Missing Fields: the LLM may not understand the complete set of arguments necessary for a given API call
     \item Hallucinated Fields: the LLM may imagine additional fields that don't exist -- a long noted trait of LLMs \citep{goddard2023hallucinations}.
 \end{enumerate}
 
Such agents may also exhibit \textbf{general planning errors}, in which they suggest an incorrect order or set of actions to partake in, thereby limiting their effectiveness.
 
As we will argue further on, a critical tension arises between skills related to in-context learning (ICL) and memorization: we will argue that the parsing errors may be addressed through an agent skilled in memorizing, whereas the planning errors may be addressed with an agent skilled in in-context learning, and improvement in one area necessarily leads to decline in the other. 

To address these challenges, we propose a novel factored agent architecture that decouples the roles of memory and context adaptation. Our design partitions the agent into two specialized components: 

\begin{enumerate}
    \item A larger in-context learner that dynamically assimilates new information from the prompt and plans an appropriate tool-use prompt.
    \item A small language model which acts as a memorizer for tool APIs that retains and retrieves long-term knowledge.
\end{enumerate}

Our factored agent thereby mitigates the drawbacks of the traditional single-agent approach while retaining its core benefits.

The contributions of this paper are as follows:

\begin{itemize}
\item We introduce a factored agent architecture that distinctly separates memorization and in-context learning.
\item We provide a detailed critique of current single-agent frameworks, highlighting issues such as malformed, missing, and hallucinated fields, as well as general planning errors.
\item We analyze the trade-offs between in-context learning and static memorization, offering insights into their respective advantages and limitations.
\item We demonstrate through empirical evaluation that our proposed architecture significantly improves planning accuracy and error resilience in real-world tasks.
\end{itemize}

In the subsequent sections, we elaborate on the design and implementation of our factored agent, present comprehensive experimental results, and discuss the broader implications of our work for the future of agentic AI.

\section{Background Literature}

Agentic AI is a rapidly evolving field that seeks to create intelligent agents capable of performing tasks requiring both in-context learning (ICL) and memorization. A central challenge in this endeavor is that within modern single agents, there appears to be a trade off between these skills. Recent work, such as \cite{iclvsmemorization}, observes that as an agent becomes more of a memorizer, its ICL capabilities diminish, and vice versa; \cite{biderman2024loralearnsforgets} report this finding as well in Low Rank Adaptor models. Regrettably for tool use models, memorization tends to increase in larger models \citep{satvaty2024undesirablememorizationlargelanguage, kiyomaru-etal-2024-comprehensive-analysis} and seems to occur in current tool use agents \citep{toolformer}. This suggests that a single agent may not be optimal for all tasks.

To address this, researchers have explored various approaches. TinyAgents \citep{erdogan2024tinyagentfunctioncallingedge} fine-tune small language models for specific tasks, but this approach may not scale well to a large number of tasks. ToolFormer \citep{schick2023toolformer} is a transformer specifically trained to generate API calls, but it may hallucinate in novel situations. \cite{qin2024toolllm} trained ToolLLaMA on over 16,000 APIs (composing the ToolBench dataset) and evaluated on a subset. Gorilla \citep{patil2023gorillalargelanguagemodel} is a more general agent that can interact with APIs by including API documentation in a Retrieval Augmented Generation (RAG) approach, and maintains a continually expanding database of API calls. However, both approaches may fail when performing supervised fine-tuning (SFT) on the same model that may also need planning and ICL abilities.

A promising direction is to combine the strengths of memorization and ICL. ReAct \citep{yao2023reactsynergizingreasoningacting} is an example of an agent that reasons and acts in the world. ReAct focuses on general orchestration of two agents, with the ``reasoning'' agent being a base reasoner, and the ``acting'' agent being the tool agent, but they often rely on external tools and knowledge bases. Other, older approaches have factored agent architectures: \cite{wen2016network} and \cite{wu2019globaltolocalmemorypointernetworks} are notable examples. 

By carefully balancing memorization and ICL, we can create more versatile and intelligent agents.

\section{System Design}

\begin{figure*}
    \centering
    \includegraphics[width=.9\textwidth]{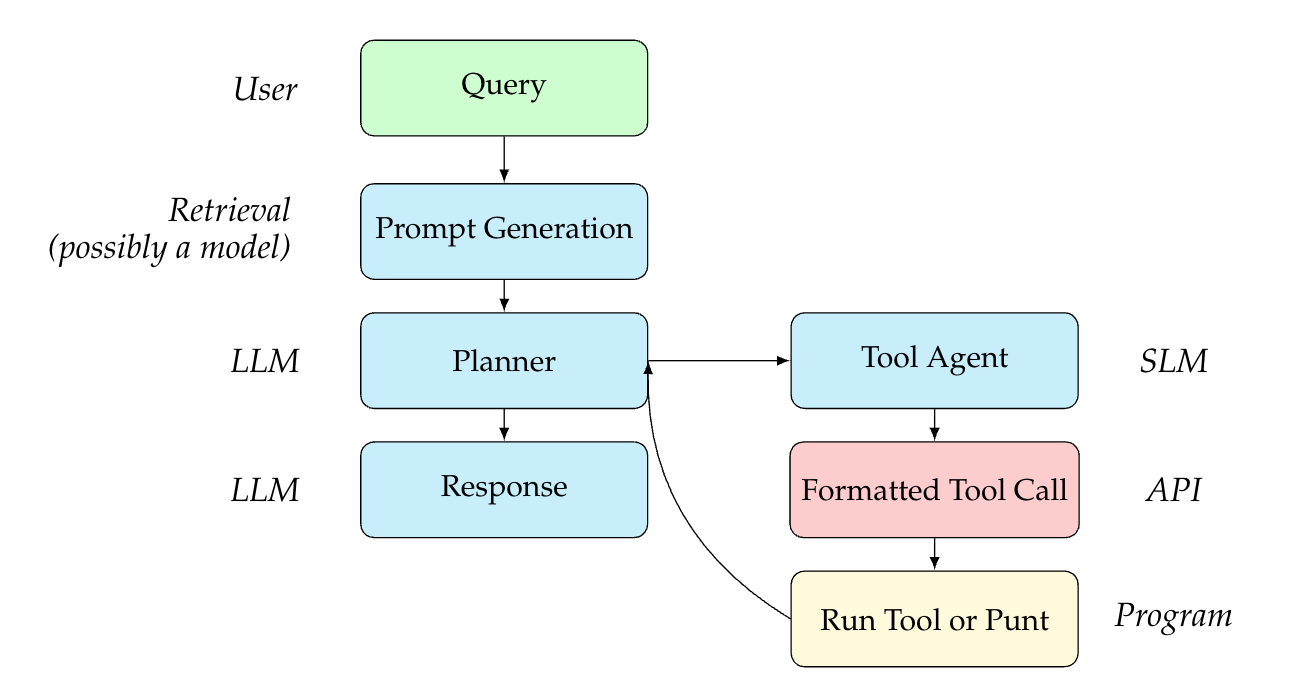}
    \caption{Representation of the factored agents model with a flow of information between users, with hand-offs between various models.}
    \label{fig:model-diagram}
\end{figure*}

Please see Figure \ref{fig:model-diagram} for a diagram of the model setup and a flow of intermediate hand-offs between agents in our factored approach. 

In an abstracted example, let's assume that the user may desire to adopt a cat in San Francisco. They would initiate an interaction with the system, i.e. inputting a query like \textit{``Book the first available interview with Paws Cat Shelter in San Francisco.''} Such an interaction would make an API call to the calendar of Paws Cat Shelter. 

In our setup, the query is first handled by the prompt generator, which may be an LLM or may be as simple as an automatic prompt augmenter, concatenating information about the person or setting to the query (i.e. demographic information that Paws Cat Shelter may need to know). This is passed to a planning agent, an LLM, that chooses which API to engage with and includes the appropriate information given the type of API (date, time, etc.). This LLM then passes a query to the Tool Agent, a Small Language Model (SLM) whose job it is to format the tool call. The query is a natural language representation of the tool call, akin to a chain of thought or a summary, e.g. \textit{``Use the paws\_shelter tool to book an interview on November 2, 2025.''}, whereas the formatted tool call returned by the SLM will have the proper syntax (e.g. \textit{``\{"api": "paws\_shelter", "operation\_id": "book\_interview", "date": "2025-11-02"\}''}). After the tool call is used on the API, the API response is returned to the Planner, which then invokes its underlying program that generates a detailed prompt capturing the necessary details (e.g., name, date, type of appointment, adoption context, etc.).

Our planner is a much larger agent, which is necessary to devise an efficient and possibly creative plan. The tool agent, meanwhile, is small and fine-tuned to adjust to specific classes of APIs. We will show ablations around this in the discussion. 
\section{Experimental Setup}

\subsection{Model Configurations}

For each benchmark, we train a tool-agent on a synthetic dataset created from the tools exposed in the training corpus. Thus, we have different tool-agents for each benchmark. We use an off-the-shelf LLM (GPT-4o) as a planner agent to coordinate the tool-use by dispatching natural language calls to the tool-agent. The output of the tool-agent is dispatched to the test tool/API.

We use Gemini Pro 1.5 \citep{geminiteam2024gemini15unlockingmultimodal} to create a synthetic dataset given only the tool schemas in the training data (effectively a zero-shot approach). As a pre-processing step, we convert the schemas to OpenAPI format.
First, because previous work indicated that diverse personas also result in better synthetic data \citep{ge2024scalingsyntheticdatacreation}, we uniformly sample a persona from a predefined list of names and locations.\footnote{https://github.com/dominictarr/random-name} Given the tool schema, persona, and a Python environment with the $numpy$ library, we instruct the model to write Python code to create a world state, which may include random dates, for example.
Then, we instruct the model to generate conversations conditioned on the world state and a uniformly-sampled number of user/assistant turns.
At the next stage, we generate a ``natural language tool call'' given the output of the previous stages.
Finally, we generate a structured tool call (as JSON) and only accept candidates where the model judges that the JSON tool call is faithful to the context (conversation and natural language tool call).
At each stage, we accept the first response from the LLM that satisfies all criteria. If we reject all candidates at a given stage, we exit early and return no synthetic example.

\subsection{Benchmarks}

\subsubsection{Tool-Use Benchmarks}
We tested our models on 
\textbf{TauBench} \citep{yao2024tau}. TauBench is a benchmark designed to evaluate the performance and reliability of tool-use agents. TauBench tests agents on multi-turn dialogues where an AI agent must dynamically interact with a simulated human user and various domain-specific tools, specifically focusing on those used in retail and airline scenarios. It simulates realistic databases, APIs, and policy guidelines. Additionally, TauBench defines and uses a multi-trial metric, ``Pass\textasciicircum \textit{k}'', which tests an agent multiple times on the same task. All models, including our own, struggle to output the same function calls on multiple passes, and so pass \textasciicircum \textit{k} metrics for baselines reported by TauBench steadily decrease as \textit{k} increases.


\subsubsection{In-context learning benchmarks}

We use two datasets to test in-context learning for a discussion on mechanistic interpretability: \textbf{BigBenchHard} (BBH) and \textbf{Grade School Math 8k (questions)} (GSM8k). BigBenchHard (BBH) is a curated subset of the larger BIG-bench benchmark, which was originally developed by Google along with contributions from the research community. BBH specifically comprises a set of challenging tasks, including boolean expressions, logical deduction with varying numbers of objects, temporal sequences, and more abstract challenges like hyperbaton and disambiguation QA.  Prior work has leaned on BBH as harder problems requiring multi-step in context learning \citep{agarwal2024many} as many questions are designed specifically to be unsearchable   \citep{cobbe2021trainingverifierssolvemath}.

GSM8k is a collection of grade school math problems, including addition, subtraction, multiplication, and division, as well as more complex tasks like multi-step reasoning, unit conversion, and basic probability. Many of the problems are framed as word problems that require careful reading, extraction of relevant information, and sequential problem solving. We claim, as others have \citep{wei2022chain}, that \textit{GSM8K requires inductive skills} due to math problems being extrapolations from other forms given prior insight. This is closer to the type of in-context learning from papers we hypothesis on. These are questions that clearly aren't searchable and unlikely memorizable because they are like solutions to math questions. There are too many variations to enumerate and store.

We chose these benchmarks motivated by mechanistic interpretability research from Anthropic \citep{iclvsmemorization} and Google Deepmind \citep{chan2022data}. Both groups demonstrate the presence of induction circuits (i.e. conditional copying of data between semantic locations for ICL) in small neural networks and their function in simple in-context learning tasks. The latter work also explores how in-context learning ability diminishes overall as a model memorizes out-of-distribution (OOD) information, despite this causing the model to perform better within the previously-OOD data specifically.

\subsection{Training Setup}

For tool-agent model training, we use Hugging Face TRL \citep{vonwerra2022trl} and transformers \citep{wolf-etal-2020-transformers} libraries for supervised fine-tuning, leveraging DeepSpeed \citep{rajbhandari2020zeromemoryoptimizationstraining} through accelerate \citep{accelerate}.

The synthetic assistant-style conversational data are used to condition the tool-agent before generating the structured tool-call request.

For TauBench, we use 11,241 synthetic examples to train on 30 tools -- 14 airline tools and 16 retail tools. 

%

\subsection{Baseline}

We choose to compare our method to a base model with a specific prompt suffix that encourages the model to output JSON. The suffix is listed in Appendix \ref{sec:baseline_suffix}. 

Comparing our model to the state of the art GPT-4o Function Calling (FC) agent would not be an altogether fair baseline: GPT-4o FC has many other optimizations that are orthogonal to and may be implemented alongside our factored agents approach. Thus, although we considered using GPT-4o FC as a baseline, we ultimately decided to use a more apples-to-apples comparison instead.

\subsection{Computational Setup}

Tool-agent training was conducted using 2x to 8x A100 NVIDIA GPUs per experiment.

\subsection{Ablations and details}

To compare the tool-agent's ability to generate well-formed requests with correct tool and parameters, we utilize GPT-4o as a planner to convert the high level request into individual natural language tool-requests. We ablate on the LLM architecture used, cycling between Gemma 3 \citep{gemma_2025} (1.5B, 4B, 12B) and DeepSeek R1 distilled Qwen \citep{deepseekai2025deepseekr1incentivizingreasoningcapability} (1.5B and 7B) to compare impact of base model and size. These models were carefully chosen to have been less likely to include TauBench training data in their pretraining corpus.

\subsection{In Context Learning}

In order to understand the impact of training structured tool-call generation on natural language models, we run BIG-Bench Hard (BBH) \citep{srivastava2023beyond} on the tool-agents to measure the impact of structured tool-use on general in-context-learning ability. BBH has frequently been used in the literature for testing ICL \citep{agarwal2024many}.
\section{Results}

Table 1 presents the pass-hat-k accuracy (P\textasciicircum1 through P\textasciicircum4) for the Retail and Airline test series. In the Retail domain, the baseline model (GPT‐4o + FAEP) achieves lower performance than each of the Gemma3 configurations, with the 4 billion‐parameter variant obtaining the highest values overall (0.45 at P\textasciicircum1 and 0.23 at P\textasciicircum4). The DeepSeek Qwen Distill models also outperform the baseline on Retail, though their scores (e.g., 0.44 at P\textasciicircum1 for the 7B version) remain slightly below Gemma 4b at the higher passes. This pattern suggests that funneling additional parameters or using more advanced tool‐use approaches can yield consistent gains for Retail tasks.

Conversely, on the Airline tests, only the Gemma3 12b and the DeepSeek1.5B models exceed baseline performance at P\textasciicircum1, with Gemma3 12b producing the top scores across all passes (e.g., 0.31 at P\textasciicircum1, 0.16 at P\textasciicircum4). Interestingly, Gemma3 1b and 4b drop below the baseline’s accuracy, indicating that merely adding parameters may not guarantee improvements in this domain. Overall, the results highlight variation in how different tool‐use or parameter expansions translate into accuracy gains, depending on the task domain and pass depth.

\begin{table*}[ht]
\centering
\caption{Side-by-side results for Retail and Airline test series. Please note that the baseline, FAEP, is the Factored Agents Equivalent Prompt. All $P\textasciicircum k$ are abbreviations of Pass$\textasciicircum k$. All rows are defined by a GPT4o base model with an additional tool-use model $i$ (defined in "+ $i$").}
\label{tab:retail-airline}
\begin{tabular}{lcccccccc}
\toprule
& \multicolumn{4}{c}{\textbf{Retail}} & \multicolumn{4}{c}{\textbf{Airline}} \\
\cmidrule(lr){2-5} \cmidrule(lr){6-9}
\textbf{Model: GPT4o +} & \textbf{P$\textasciicircum 1$} & \textbf{P$\textasciicircum 2$} & \textbf{P$\textasciicircum 3$} & \textbf{P$\textasciicircum 4$} & \textbf{P$\textasciicircum 1$} & \textbf{P$\textasciicircum 2$} & \textbf{P$\textasciicircum 3$} & \textbf{P$\textasciicircum 4$} \\
\midrule
+ FAEP (baseline)
    & 0.37 & 0.23 & 0.17 & 0.14
    & 0.29 & 0.19 & 0.15 & 0.12 \\
+ Gemma3 1b
    & 0.41 & 0.26 & 0.19 & 0.16
    & 0.22 & 0.15 & 0.11 & 0.08 \\
+ Gemma3 4b
    &\textbf{ 0.45} & \textbf{0.33} & \textbf{0.27} & \textbf{0.23 }
    & 0.25 & 0.17 & 0.14 & 0.12 \\
+ Gemma3 12b
    & 0.42 & 0.30 & 0.24 & 0.21
    & \textbf{0.31} & \textbf{0.23 }& \textbf{0.19} & \textbf{0.16} \\
+ DeepSeek Qwen Distill 1.5B
    & 0.39 & 0.28 & 0.23 & 0.21
    & 0.31 & 0.20 & 0.16 & 0.12 \\
 + DeepSeek Qwen Distill 7B
    & 0.44 & 0.31 & 0.24 & 0.20
    & 0.26 & 0.18 & 0.13 & 0.10 \\
\bottomrule
\end{tabular}
\end{table*}

\section{Discussion}

\begin{figure*}
    \centering
    \includegraphics[width=.95\textwidth]{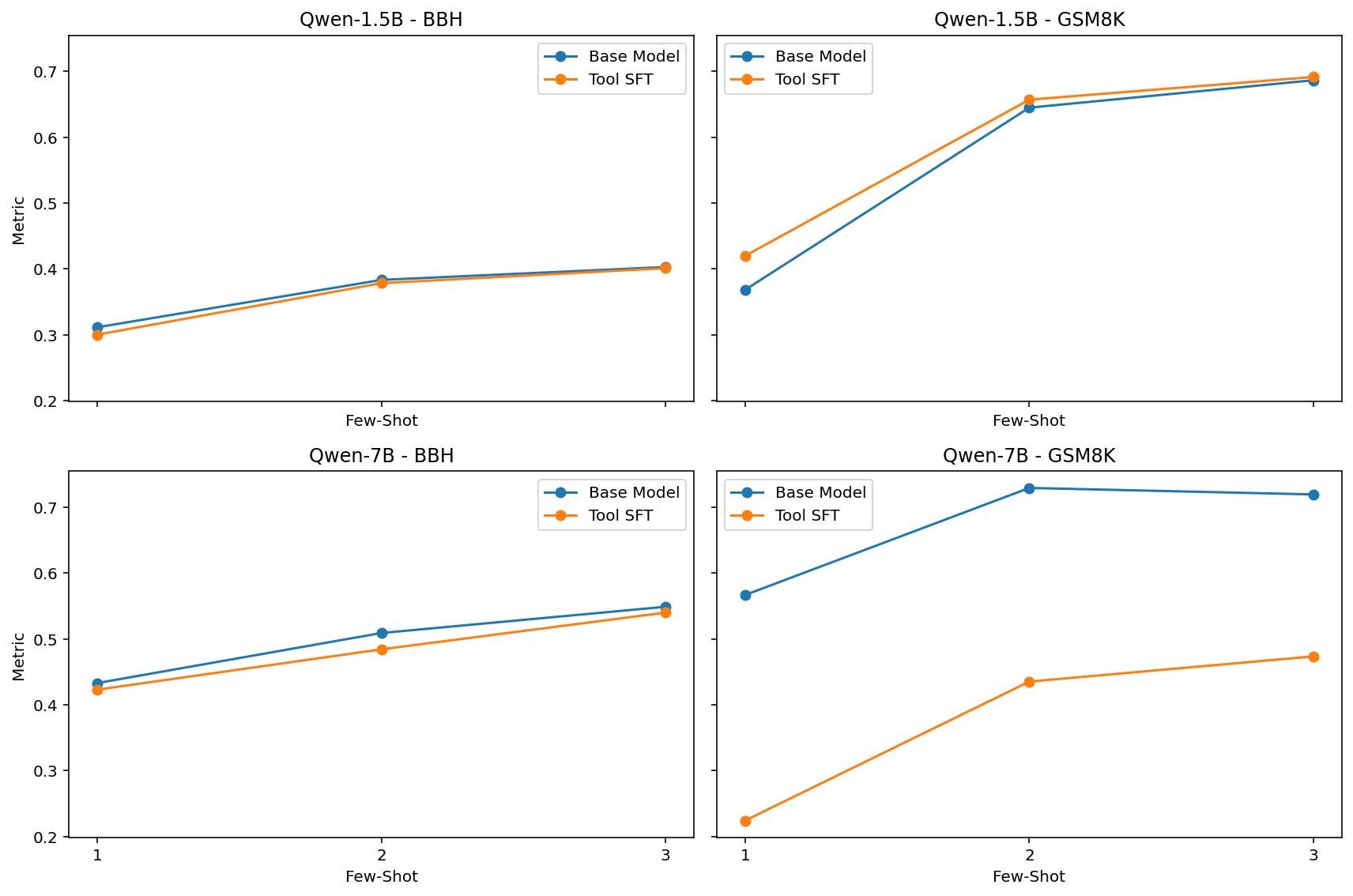}
    \caption{In this plot, we compare the DeepSeek-R1-Distill-Qwen base model to the tool-use model across two benchmarks, BigBenchHard and GSM8k. The x-axis increases the number of shots present in few-shot test time, and the y axis measures proportion of scored answers that are an exact match.}
    \label{fig:plot1}
\end{figure*}

The two main hypotheses we seek to evaluate are:
\begin{enumerate}
    \item That in-context learning trades off with memorization of tool semantics, and
    \item That using separate models for in-context learning tasks from memorizing tools can improve performance on tool-use tasks
\end{enumerate}

We wanted to explore this work with simple experiments geared towards a practical tool-use context. It is common practice to memorize tools (i.e. ToolLlama, Toolformer, Gorilla) in-weights, which seems at odds with the basic findings of this body of prior work. If we see a degradation in in-context learning and an improvement in performance from memorizing specific tools with a separate model, it will support the existence of this tradeoff in a highly-relevant, practical setting while also demonstrating that the tradeoff can be avoided, at least in part, by separating these fundamental responsibilities into separate models. Future work may further extend these experiments, examining how the performance improvements from Factored Agents stack with techniques like tool RAG, chain-of-thought prompting, ReAct, Act, constrained decoding, reinforcement learning (i.e. RL on natural\_language\_tool\_call and tool environments more generally), and other commonly-applied planning and tool-use methods.


Our performance on TauBench of a baseline Factored Agents implementation with a naive prompt-only implementation shows what the prior work leads us to expect -- that is, not only from a larger model do abilities like in-context learning emerge, but ~7B is a reasonable inflection point of such properties, especially given more modern training techniques (see Figure 4 in \cite{wei2022chain}.) We see a performance benefit by sampling from a model that has been fine-tuned to memorize the tools we use rather than sampling from GPT-4o, which attempts to model a much wider distribution of knowledge. In both cases, the output is conditioned on a very similar, simple prompt.

Our investigation into the degradation of in-context learning abilities under out-of-distribution (OOD) fine-tuning yielded unexpected findings. Consistent with the observations reported in the Google DeepMind study, our results indicate that in-context learning performance deteriorates following OOD fine-tuning, as evidenced by a significant decline on the GSM8K benchmark. In contrast, \textit{the performance on BigBenchHard does not exhibit a similarly pronounced degradation}, suggesting a differential sensitivity to OOD fine-tuning across these benchmarks.

We claim, as others have, that GSM8K requires inductive skills due to math problems being extrapolations from other forms given prior insight. This is closer to the type of in-context learning from the papers that prompted our work. These are questions that clearly are not searchable and are unlikely memorizable because, as solutions to math questions, they populate a space too large to enumerate and store.

In contrast to GSM8K, BigBenchHard contains tasks like: (1) logical deduction between three objects, (2) five objects, and (3) seven objects, (4) sports understanding, (5) ruin names, (6) object counting, (7) disambiguation qa, and (8) snarks that rely heavily on process that are more difficult to frame as the simple logical induction tasks that these papers study. It is possible that \textit{many different learning mechanisms} beyond in-context learning are required for these tasks, complicating their interpretation.

The plots for the small 1.5B-parameter model all show no change between the finetuned model and the base model on our selected in-context learning benchmarks. This is also expected given the prior work \citep{wei2022chain} shows how in-context learning performance increases dramatically as models increase in size. Such a small model is likely to resort to memorization-oriented mechanisms disproportionately, and thus we can expect such models to be less-effected in this way by memorization-oriented training, though we might expect to see a small performance decrease due to catastrophic forgetting, which makes these results still slightly surprising.





\section{Conclusion}

Our study demonstrates that decoupling the roles of memorization and context adaptation within an agentic AI framework leads to marked improvements in standard tool-use benchmarks, implying an improvement of planning and reduction of errors. The proposed factored agent architecture outperforms traditional single-agent designs by specifically addressing issues such as incorrect API output and flawed planning sequences. Through ablations on model size and standard ICL benchmarks, we have shown that separating the memory and in-context learning functionalities not only enhances overall system robustness but also provides a flexible platform for integrating advanced tool-use strategies. These results underscore the potential of factored agents to serve as a foundational paradigm for next-generation agentic AI, while also highlighting avenues for future research on scalability and the integration of additional specialized modules.

\section{Limitations and Future Work}

Our work \textbf{has a number of limitations.} We include a limited number of benchmarks relative to the breadth of those included in other work \citep{wei2022chain, agarwal2024many}. Our setup only entails a unidirectional pathway between planner and tool user -- if the planner could call the tool agent multiple times or engage in a dialogue, the output may lead to more general tool calling. 




\textbf{In the future}, we hope to expand our analysis to improve the error resilience of our models. First, testing with additional methods on the planner model, like Low-Rank Adaptation (LoRA) and reinforcement learning, that could elicit stronger planning performance with natural-language tool-calls. We hope to experiment with Classifier Free Guidance (CFG) as a method to direct the planner model to focus on relevant parts of the prompt, perhaps improving the relevance of the information observed. As prior work has shown that this can reduce the prevalence of hallucinations \citep{sanchez2023staytopicclassifierfreeguidance}, we believe that this may force the LLM to focus on the natural language tool call. We would also like to benchmark factored agents as part of a more sophisticated tool-use system, incorporating advanced prompting techniques and inference-time optimizations like constrained decoding.

Finally, we would like to focus model's effort on the rest of the conversation, tending to errors by using their specialized understanding of the tool's semantics. Techniques like dropout and noising natural-language tool-calls when training tool agents could lead them to catch additional errors and hallucinations made by the planner.

\section{Ethical Considerations}

Here, we consider possible ethical ramifictions of our work. Memorization of tool-related data may, if trained improperly, may result in recitation of unwanted details. We do not provide this training ourselves, and we would advise that those specifically worried about this outcome may consider using the smallest tool-use models possible at the expense of degradation of services. 

Better tool-use models may be used to automate the use of unethical tools or may be used in unethical tool-use system. Our work does not directly provide this access.  

\bibliography{custom}

\begin{thebibliography}{29}
\providecommand{\natexlab}[1]{#1}
\providecommand{\url}[1]{\texttt{#1}}
\expandafter\ifx\csname urlstyle\endcsname\relax
  \providecommand{\doi}[1]{doi: #1}\else
  \providecommand{\doi}{doi: \begingroup \urlstyle{rm}\Url}\fi

\bibitem[Acharya et~al.(2025)Acharya, Kuppan, and Divya]{acharya2025agentic}
Deepak~Bhaskar Acharya, Karthigeyan Kuppan, and B~Divya.
\newblock Agentic ai: Autonomous intelligence for complex goals--a
  comprehensive survey.
\newblock \emph{IEEE Access}, 2025.

\bibitem[Agarwal et~al.(2024)Agarwal, Singh, Zhang, Bohnet, Rosias, Chan,
  Zhang, Anand, Abbas, Nova, et~al.]{agarwal2024many}
Rishabh Agarwal, Avi Singh, Lei Zhang, Bernd Bohnet, Luis Rosias, Stephanie
  Chan, Biao Zhang, Ankesh Anand, Zaheer Abbas, Azade Nova, et~al.
\newblock Many-shot in-context learning.
\newblock \emph{Advances in Neural Information Processing Systems},
  37:\penalty0 76930--76966, 2024.

\bibitem[Biderman et~al.(2024)Biderman, Portes, Ortiz, Paul, Greengard,
  Jennings, King, Havens, Chiley, Frankle, Blakeney, and
  Cunningham]{biderman2024loralearnsforgets}
Dan Biderman, Jacob Portes, Jose Javier~Gonzalez Ortiz, Mansheej Paul, Philip
  Greengard, Connor Jennings, Daniel King, Sam Havens, Vitaliy Chiley, Jonathan
  Frankle, Cody Blakeney, and John~P. Cunningham.
\newblock Lora learns less and forgets less, 2024.
\newblock URL \url{https://arxiv.org/abs/2405.09673}.

\bibitem[Chan et~al.(2022)Chan, Santoro, Lampinen, Wang, Singh, Richemond,
  McClelland, and Hill]{chan2022data}
Stephanie Chan, Adam Santoro, Andrew Lampinen, Jane Wang, Aaditya Singh, Pierre
  Richemond, James McClelland, and Felix Hill.
\newblock Data distributional properties drive emergent in-context learning in
  transformers.
\newblock \emph{Advances in neural information processing systems},
  35:\penalty0 18878--18891, 2022.

\bibitem[Cobbe et~al.(2021)Cobbe, Kosaraju, Bavarian, Chen, Jun, Kaiser,
  Plappert, Tworek, Hilton, Nakano, Hesse, and
  Schulman]{cobbe2021trainingverifierssolvemath}
Karl Cobbe, Vineet Kosaraju, Mohammad Bavarian, Mark Chen, Heewoo Jun, Lukasz
  Kaiser, Matthias Plappert, Jerry Tworek, Jacob Hilton, Reiichiro Nakano,
  Christopher Hesse, and John Schulman.
\newblock Training verifiers to solve math word problems, 2021.
\newblock URL \url{https://arxiv.org/abs/2110.14168}.

\bibitem[DeepSeek-AI(2025)]{deepseekai2025deepseekr1incentivizingreasoningcapability}
DeepSeek-AI.
\newblock Deepseek-r1: Incentivizing reasoning capability in llms via
  reinforcement learning, 2025.
\newblock URL \url{https://arxiv.org/abs/2501.12948}.

\bibitem[Erdogan et~al.(2024)Erdogan, Lee, Jha, Kim, Tabrizi, Moon, Hooper,
  Anumanchipalli, Keutzer, and
  Gholami]{erdogan2024tinyagentfunctioncallingedge}
Lutfi~Eren Erdogan, Nicholas Lee, Siddharth Jha, Sehoon Kim, Ryan Tabrizi,
  Suhong Moon, Coleman Hooper, Gopala Anumanchipalli, Kurt Keutzer, and Amir
  Gholami.
\newblock Tinyagent: Function calling at the edge, 2024.
\newblock URL \url{https://arxiv.org/abs/2409.00608}.

\bibitem[Ge et~al.(2024)Ge, Chan, Wang, Yu, Mi, and
  Yu]{ge2024scalingsyntheticdatacreation}
Tao Ge, Xin Chan, Xiaoyang Wang, Dian Yu, Haitao Mi, and Dong Yu.
\newblock Scaling synthetic data creation with 1,000,000,000 personas, 2024.
\newblock URL \url{https://arxiv.org/abs/2406.20094}.

\bibitem[Georgiev et~al.(2024)Georgiev, Lei, Burnell, Bai, Gulati, Tanzer,
  Vincent, Pan, Wang, Mariooryad, Ding, Geng, Alcober, Frostig, Omernick,
  Walker, Paduraru, Sorokin, Tacchetti, ..., and
  Vinyals]{geminiteam2024gemini15unlockingmultimodal}
Petko Georgiev, Ving~Ian Lei, Ryan Burnell, Libin Bai, Anmol Gulati, Garrett
  Tanzer, Damien Vincent, Zhufeng Pan, Shibo Wang, Soroosh Mariooryad, Yifan
  Ding, Xinyang Geng, Fred Alcober, Roy Frostig, Mark Omernick, Lexi Walker,
  Cosmin Paduraru, Christina Sorokin, Andrea Tacchetti, ..., and Oriol Vinyals.
\newblock Gemini 1.5: Unlocking multimodal understanding across millions of
  tokens of context, 2024.
\newblock URL \url{https://arxiv.org/abs/2403.05530}.

\bibitem[Goddard(2023)]{goddard2023hallucinations}
Jerome Goddard.
\newblock Hallucinations in chatgpt: A cautionary study.
\newblock \emph{Journal of Biomedical Informatics}, 45:\penalty0 123--130,
  2023.

\bibitem[Gugger et~al.(2022)Gugger, Debut, Wolf, Schmid, Mueller, Mangrulkar,
  Sun, and Bossan]{accelerate}
Sylvain Gugger, Lysandre Debut, Thomas Wolf, Philipp Schmid, Zachary Mueller,
  Sourab Mangrulkar, Marc Sun, and Benjamin Bossan.
\newblock Accelerate: Training and inference at scale made simple, efficient
  and adaptable., 2022.
\newblock URL \url{https://github.com/huggingface/accelerate}.

\bibitem[Kamath et~al.(2025)Kamath, Ferret, Pathak, Vieillard, Merhej, Perrin,
  Matejovicova, Ramé, Rivière, Rouillard, Mesnard, Cideron, bastien Grill,
  Ramos, Yvinec, Casbon, Pot, Penchev, Liu, ..., and Hussenot]{gemma_2025}
Aishwarya Kamath, Johan Ferret, Shreya Pathak, Nino Vieillard, Ramona Merhej,
  Sarah Perrin, Tatiana Matejovicova, Alexandre Ramé, Morgane Rivière, Louis
  Rouillard, Thomas Mesnard, Geoffrey Cideron, Jean bastien Grill, Sabela
  Ramos, Edouard Yvinec, Michelle Casbon, Etienne Pot, Ivo Penchev, Gaël Liu,
  ..., and Léonard Hussenot.
\newblock Gemma 3.
\newblock 2025.
\newblock URL \url{https://goo.gle/Gemma3Report}.

\bibitem[Kiyomaru et~al.(2024)Kiyomaru, Sugiura, Kawahara, and
  Kurohashi]{kiyomaru-etal-2024-comprehensive-analysis}
Hirokazu Kiyomaru, Issa Sugiura, Daisuke Kawahara, and Sadao Kurohashi.
\newblock A comprehensive analysis of memorization in large language models.
\newblock In Saad Mahamood, Nguyen~Le Minh, and Daphne Ippolito (eds.),
  \emph{Proceedings of the 17th International Natural Language Generation
  Conference}, pp.\  584--596, Tokyo, Japan, September 2024. Association for
  Computational Linguistics.
\newblock URL \url{https://aclanthology.org/2024.inlg-main.45/}.

\bibitem[Li et~al.(2023{\natexlab{a}})Li, Li, Li, and Zhang]{iclvsmemorization}
Yiding Li, Haotian Li, Jianwei Li, and Tong Zhang.
\newblock Icl heads and data distributional heads drive in context learning.
\newblock \emph{arXiv preprint arXiv:2305.16880}, 2023{\natexlab{a}}.

\bibitem[Li et~al.(2023{\natexlab{b}})Li, Li, Li, and Zhang]{toolformer}
Yuchen Li, Haotian Li, Jianwei Li, and Tong Zhang.
\newblock Toolformer: Interactive learning of tool use from human feedback.
\newblock In \emph{NeurIPS}, 2023{\natexlab{b}}.

\bibitem[Patil et~al.(2023)Patil, Zhang, Wang, and
  Gonzalez]{patil2023gorillalargelanguagemodel}
Shishir~G. Patil, Tianjun Zhang, Xin Wang, and Joseph~E. Gonzalez.
\newblock Gorilla: Large language model connected with massive apis, 2023.
\newblock URL \url{https://arxiv.org/abs/2305.15334}.

\bibitem[Qin et~al.(2024)Qin, Liang, Ye, Zhu, Yan, Lu, Lin, Cong, Tang, Qian,
  Zhao, Hong, Tian, Xie, Zhou, Gerstein, dahai li, Liu, and
  Sun]{qin2024toolllm}
Yujia Qin, Shihao Liang, Yining Ye, Kunlun Zhu, Lan Yan, Yaxi Lu, Yankai Lin,
  Xin Cong, Xiangru Tang, Bill Qian, Sihan Zhao, Lauren Hong, Runchu Tian,
  Ruobing Xie, Jie Zhou, Mark Gerstein, dahai li, Zhiyuan Liu, and Maosong Sun.
\newblock Tool{LLM}: Facilitating large language models to master 16000+
  real-world {API}s.
\newblock In \emph{The Twelfth International Conference on Learning
  Representations}, 2024.
\newblock URL \url{https://openreview.net/forum?id=dHng2O0Jjr}.

\bibitem[Rajbhandari et~al.(2020)Rajbhandari, Rasley, Ruwase, and
  He]{rajbhandari2020zeromemoryoptimizationstraining}
Samyam Rajbhandari, Jeff Rasley, Olatunji Ruwase, and Yuxiong He.
\newblock Zero: Memory optimizations toward training trillion parameter models,
  2020.
\newblock URL \url{https://arxiv.org/abs/1910.02054}.

\bibitem[Sanchez et~al.(2023)Sanchez, Fan, Spangher, Levi, Ammanamanchi, and
  Biderman]{sanchez2023staytopicclassifierfreeguidance}
Guillaume Sanchez, Honglu Fan, Alexander Spangher, Elad Levi, Pawan~Sasanka
  Ammanamanchi, and Stella Biderman.
\newblock Stay on topic with classifier-free guidance, 2023.
\newblock URL \url{https://arxiv.org/abs/2306.17806}.

\bibitem[Satvaty et~al.(2024)Satvaty, Verberne, and
  Turkmen]{satvaty2024undesirablememorizationlargelanguage}
Ali Satvaty, Suzan Verberne, and Fatih Turkmen.
\newblock Undesirable memorization in large language models: A survey, 2024.
\newblock URL \url{https://arxiv.org/abs/2410.02650}.

\bibitem[Schick et~al.(2023)Schick, Dwivedi-Yu, Dess{\`\i}, Raileanu, Lomeli,
  Hambro, Zettlemoyer, Cancedda, and Scialom]{schick2023toolformer}
Timo Schick, Jane Dwivedi-Yu, Roberto Dess{\`\i}, Roberta Raileanu, Maria
  Lomeli, Eric Hambro, Luke Zettlemoyer, Nicola Cancedda, and Thomas Scialom.
\newblock Toolformer: Language models can teach themselves to use tools.
\newblock \emph{Advances in Neural Information Processing Systems},
  36:\penalty0 68539--68551, 2023.

\bibitem[Srivastava et~al.(2023)Srivastava, Rastogi, Rao, Shoeb, Abid, Fisch,
  Brown, Santoro, Gupta, Garriga-Alonso, Kluska, Lewkowycz, Agarwal, Power,
  Ray, Warstadt, Kocurek, Safaya, Tazarv, ..., and Wu]{srivastava2023beyond}
Aarohi Srivastava, Abhinav Rastogi, Abhishek Rao, Abu Awal~Md Shoeb, Abubakar
  Abid, Adam Fisch, Adam~R. Brown, Adam Santoro, Aditya Gupta, Adrià
  Garriga-Alonso, Agnieszka Kluska, Aitor Lewkowycz, Akshat Agarwal, Alethea
  Power, Alex Ray, Alex Warstadt, Alexander~W. Kocurek, Ali Safaya, Ali Tazarv,
  ..., and Ziyi Wu.
\newblock Beyond the imitation game: Quantifying and extrapolating the
  capabilities of language models.
\newblock \emph{Transactions on Machine Learning Research}, 2023.
\newblock ISSN 2835-8856.
\newblock URL \url{https://openreview.net/forum?id=uyTL5Bvosj}.

\bibitem[von Werra et~al.(2020)von Werra, Belkada, Tunstall, Beeching, Thrush,
  Lambert, Huang, Rasul, and Gallouédec]{vonwerra2022trl}
Leandro von Werra, Younes Belkada, Lewis Tunstall, Edward Beeching, Tristan
  Thrush, Nathan Lambert, Shengyi Huang, Kashif Rasul, and Quentin Gallouédec.
\newblock Trl: Transformer reinforcement learning, 2020.
\newblock URL \url{https://github.com/huggingface/trl}.

\bibitem[Wei et~al.(2022)Wei, Wang, Schuurmans, Bosma, Chi, Le, and
  Zhou]{wei2022chain}
Jason Wei, Xuezhi Wang, Dale Schuurmans, Maarten Bosma, Ed~Chi, Quoc Le, and
  Denny Zhou.
\newblock Chain-of-thought prompting elicits reasoning in large language
  models.
\newblock \emph{arXiv preprint arXiv:2201.11903}, 2022.

\bibitem[Wen et~al.(2016)Wen, Vandyke, Mrksic, Gasic, Rojas-Barahona, Su,
  Ultes, and Young]{wen2016network}
Tsung-Hsien Wen, David Vandyke, Nikola Mrksic, Milica Gasic, Lina~M
  Rojas-Barahona, Pei-Hao Su, Stefan Ultes, and Steve Young.
\newblock A network-based end-to-end trainable task-oriented dialogue system.
\newblock \emph{arXiv preprint arXiv:1604.04562}, 2016.

\bibitem[Wolf et~al.(2020)Wolf, Debut, Sanh, Chaumond, Delangue, Moi, Cistac,
  Rault, Louf, Funtowicz, Davison, Shleifer, von Platen, Ma, Jernite, Plu, Xu,
  Scao, Gugger, Drame, Lhoest, and Rush]{wolf-etal-2020-transformers}
Thomas Wolf, Lysandre Debut, Victor Sanh, Julien Chaumond, Clement Delangue,
  Anthony Moi, Pierric Cistac, Tim Rault, Rémi Louf, Morgan Funtowicz, Joe
  Davison, Sam Shleifer, Patrick von Platen, Clara Ma, Yacine Jernite, Julien
  Plu, Canwen Xu, Teven~Le Scao, Sylvain Gugger, Mariama Drame, Quentin Lhoest,
  and Alexander~M. Rush.
\newblock Transformers: State-of-the-art natural language processing.
\newblock In \emph{Proceedings of the 2020 Conference on Empirical Methods in
  Natural Language Processing: System Demonstrations}, pp.\  38--45, Online,
  October 2020. Association for Computational Linguistics.
\newblock URL \url{https://www.aclweb.org/anthology/2020.emnlp-demos.6}.

\bibitem[Wu et~al.(2019)Wu, Socher, and
  Xiong]{wu2019globaltolocalmemorypointernetworks}
Chien-Sheng Wu, Richard Socher, and Caiming Xiong.
\newblock Global-to-local memory pointer networks for task-oriented dialogue,
  2019.
\newblock URL \url{https://arxiv.org/abs/1901.04713}.

\bibitem[Yao et~al.(2023)Yao, Zhao, Yu, Du, Shafran, Narasimhan, and
  Cao]{yao2023reactsynergizingreasoningacting}
Shunyu Yao, Jeffrey Zhao, Dian Yu, Nan Du, Izhak Shafran, Karthik Narasimhan,
  and Yuan Cao.
\newblock React: Synergizing reasoning and acting in language models, 2023.
\newblock URL \url{https://arxiv.org/abs/2210.03629}.

\bibitem[Yao et~al.(2024)Yao, Shinn, Razavi, and Narasimhan]{yao2024tau}
Shunyu Yao, Noah Shinn, Pedram Razavi, and Karthik Narasimhan.
\newblock $\tau$-bench: A benchmark for tool-agent-user interaction in
  real-world domains.
\newblock \emph{arXiv preprint arXiv:2406.12045}, 2024.

\end{thebibliography}
\bibliographystyle{colm2025_conference}

\appendix
\section{Appendix}

\subsection{Factored Agents Prompt (TauBench)}
\label{sec:fa_suffix}

\begin{quote}
Remember, to invoke a tool, output a single English paragraph, a natural language request starting with "natural\_language\_tool\_call". Only use one tool at a time.

The paragraph should include all of the tool name, parameters, and their values and nothing else. Specify the exact tool name, but the rest of the paragraph must be in English. Do not write code. Do not use JSON. Specify the exact function \/ tool name. Return the single natural\_language\_tool\_call paragraph and nothing else. For example, to call the example\_1 function, the request might be phrased as: "natural\_language\_tool\_call I will use the example\_1 function with the text argument "Hello World" to say hello."
\*\*It's very important to invoke a tool with "natural\_language\_tool\_call" in English if you say that you are going to do something, look something up, or take any action!\*\*
\end{quote}

\subsection{Factored Agents Equivalent Prompt (TauBench)}
\label{sec:baseline_suffix}

\begin{quote}
Remember, to invoke a tool, output JSON starting with "json\_tool\_call". Only use one tool at a time.

The program should include a function call with all of the tool name, parameters, and their values and nothing else. Write "json\_tool\_call", but the rest of the output must be in JSON. Do not write English. Do not use Python. Do NOT use \texttt{parameters}-- these are \texttt{arguments}. Do NOT use \texttt{functions.example\_1}-- use only \texttt{example\_1}-- never use \texttt{functions.*}! Return the single json\_tool\_call and nothing else. For example, to call the example\_1 function, the request might be phrased as: "json\_tool\_call \{"name": "example\_1", "arguments": \{"salutation": "Hello World"\}\}".

\textbf{It's very important to invoke a tool with "json\_tool\_call" if you say that you are going to do something, look something up, or take any action!}
\end{quote}

\end{document}